%% file: main.tex
\documentclass[10pt,twocolumn,letterpaper]{article}

\usepackage[pagenumbers]{iccv} 

\input{preamble}

\definecolor{iccvblue}{rgb}{0.21,0.49,0.74}
\usepackage[pagebackref,breaklinks,colorlinks,allcolors=iccvblue]{hyperref}

\def\paperID{6821} 
\def\confName{ICCV}
\def\confYear{2025}

\title{Boosting Generative Adversarial Transferability with Self-supervised\\Vision Transformer Features}

\author{Shangbo Wu$^{1}$ \quad
  Yu-an Tan$^{1}$ \quad
  Ruinan Ma$^{1}$ \quad
  Wencong Ma$^{2}$ \quad
  Dehua Zhu$^{1}$ \quad
  Yuanzhang Li$^{2}\thanks{corresponding author}$
  \\
  $ ^{1} $School of Cyberspace Science and Technology, Beijing Institute of Technology\\
  $ ^{2} $School of Computer Science and Technology, Beijing Institute of Technology\\
  {\tt\small \{shangbo.wu, tan2008, ruinan, wencong.ma, zhudehua, popular\}@bit.edu.cn}
}

\begin{document}
\maketitle
\input{sec/0_abstract}
\input{sec/1_intro}
\input{sec/2_relatedwork}
\input{sec/3_methodology}
\input{sec/4_experiments}
\input{sec/5_conclusion}

{
  \small
  \bibliographystyle{ieeenat_fullname}
  \bibliography{main}
}

\input{sec/6_suppl}
\end{document}

%% file: preamble.tex
\newcommand{\dsva}{\texttt{dSVA}}
\newcommand{\dsvab}{\textbf{\dsva}}
\newcommand{\btheta}{\bm{\theta}}
\newcommand{\bx}{\bm{x}}
\newcommand{\bxadv}{\bx^{adv}}
\newcommand{\rb}[1]{\textbf{\textcolor{red}{#1}}}

\usepackage{microtype}
\usepackage{soul}
\usepackage[T1]{fontenc}
\usepackage{bm}
\usepackage{multirow}
\usepackage{array}
\newcolumntype{*}{>{\global\let\currentrowstyle\relax}}
\newcolumntype{^}{>{\currentrowstyle}}
\newcommand{\rowstyle}[1]{\gdef\currentrowstyle{#1}%
  #1\ignorespaces
}
\renewcommand{\paragraph}[1]{
  \noindent\textbf{#1}
}

%% file: sec/0_abstract.tex
\begin{abstract}
  The ability of deep neural networks (DNNs) come from extracting and
  interpreting features from the data provided. By exploiting intermediate
  features in DNNs instead of relying on hard labels, we craft adversarial
  perturbation that generalize more effectively, boosting black-box
  transferability. These features ubiquitously come from supervised learning
  in previous work. Inspired by the exceptional synergy between
  self-supervised learning and the Transformer architecture, this paper
  explores whether exploiting self-supervised Vision Transformer (ViT)
  representations can improve adversarial transferability. We present
  \dsvab{}---a generative \underline{d}ual \underline{s}elf-supervised
  \underline{V}iT features \underline{a}ttack, that exploits both global
  structural features from contrastive learning (CL) and local textural
  features from masked image modeling (MIM), the self-supervised learning
  paradigm duo for ViTs. We design a novel generative training framework that
  incorporates a generator to create black-box adversarial examples, and
  strategies to train the generator by exploiting joint features and the
  attention mechanism of self-supervised ViTs. Our findings show that CL and
  MIM enable ViTs to attend to distinct feature tendencies, which, when
  exploited in tandem, boast great adversarial generalizability. By disrupting
  dual deep features distilled by self-supervised ViTs, we are rewarded with
  remarkable black-box transferability to models of various architectures that
  outperform state-of-the-arts. Code available at
  \href{https://github.com/spencerwooo/dSVA}{https://github.com/spencerwooo/dSVA}.
\end{abstract}

%% file: sec/1_intro.tex
\section{Introduction}
\label{sec:intro}

The transferability of adversarial examples enable real-world black-box attacks
on DNNs without the adversary's access to their internals. Such attacks require
the construction of a local white-box surrogate model. Consequently, their
effectiveness relies on the ability to disrupt the shared latent
representations, i.e., features, learnt by both models. DNNs learn sample-label
correlations over their training process, by identifying the structure and
semantic characteristics of the data for classification. These learnt deep
features are generalizable enough to essentially serve as the basis that drive
downstream tasks such as object detection~\cite{RenHG017, CarionMSUKZ20},
similarity measurement~\cite{ZhangIESW18, FuTSC0DI23}, image
super-resolution~\cite{LedigTHCCAATTWS17}, and style transfer~\cite{GatysEB16}.
Prior research has shown that improving transferability is possible by targeting
intermediate features of the surrogate model instead of directly attacking hard
labels or output gradients~\citep{WangGZLQ021, ZhouHCTHGY18, HuangKGHBL19}.
Since deep features of well-trained DNNs are generalizable~\cite{YosinskiCBL14},
perturbation designed to disrupt these features are more
transferable~\cite{InkawhichWLC19}.

\begin{figure}[t]
  \centering
  \includegraphics[width=\linewidth]{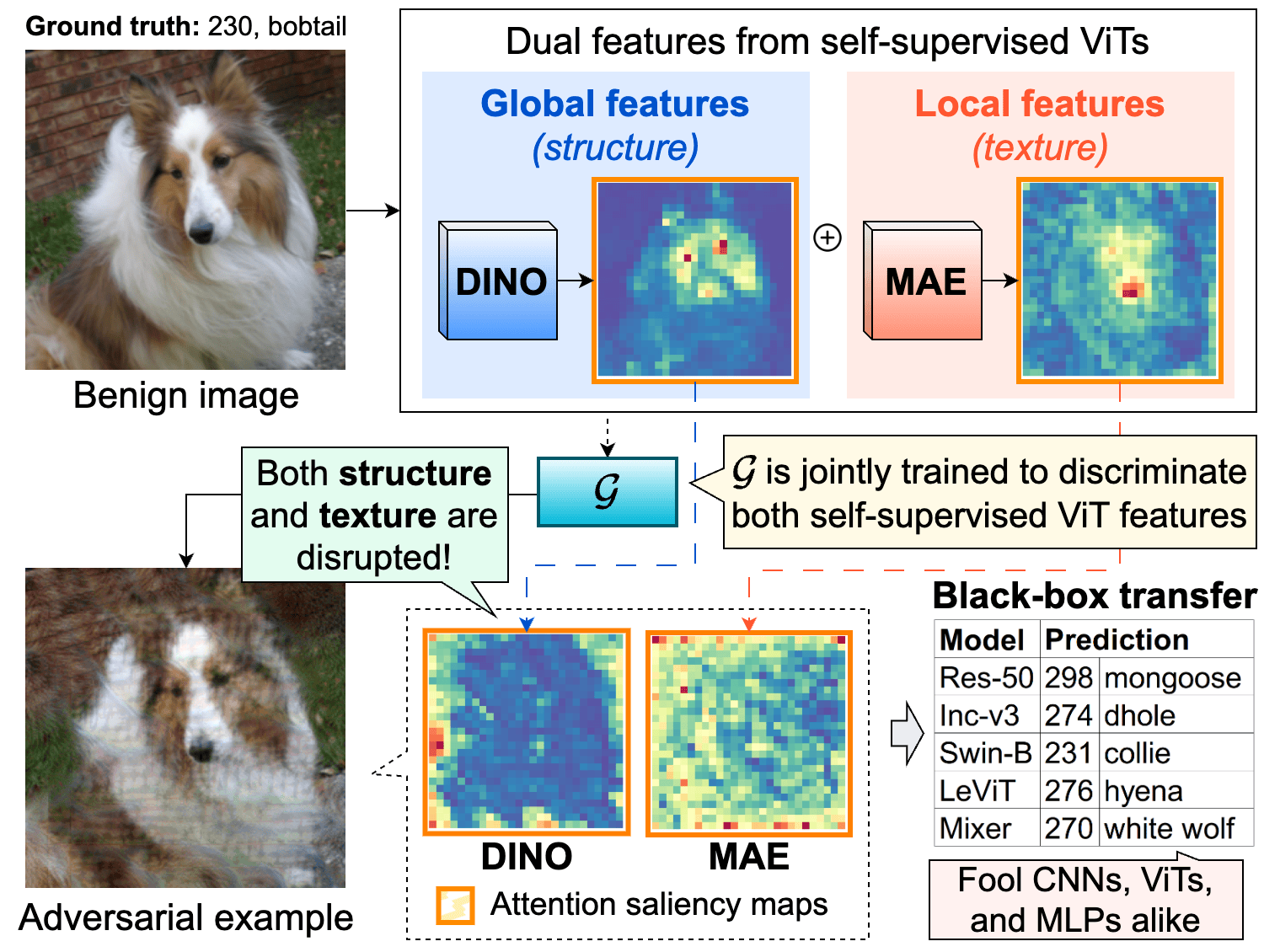}
  \caption{\textbf{Demonstration of \dsva.} By jointly exploiting deep
    features of both self-supervised ViTs, i.e., DINO (CL) and MAE (MIM),
    \dsva{} crafts perturbation that disrupts both structural and textural
    representations of the image (as visualized in the attention saliency
  maps), fooling ConvNets, ViTs, and MLPs alike.}
  \label{fig:fig1}
\end{figure}


The habitual inclusion of label-wise loss in existing work for conducting
adversarial attacks acts as a common practice that pushes the surrogate model to
be setup with supervised learning. This makes sense for ConvNets where
self-supervised learning lags behind supervised. However, the advent of ViTs
introduced the success of self-supervision in natural language processing to
vision~\cite{Bao0PW22, HeCXLDG22, ChenXH21, CaronTMJMBJ21}. Supervised learning
fails to preserve image semantics through human labelling, reducing feature-rich
semantic information within images into a single concept represented by a
human-assigned category. In contrast, self-supervised ViTs excel at capturing
semantics, providing robust positional and semantic relationships throughout
model layers, outperforming ConvNets~\cite{AmirGBD22}. Driven by the powerful
adversarial potentials of self-supervised ViT features, we ask: \textit{How can
  we fully utilize the rich representations distilled by the harmonious coalition
  between self-supervision and the Transformer architecture, to boost adversarial
transferability?} We attempt to answer this research question in threefold:

\textbf{(1) Facet-level feature exploitation.} ViTs comprise several layers of
multi-head self-attention blocks that encode token-wise features. With a goal of
extracting adversarially generalizable features, contrary to ConvNets where
existing work use the direct output of entire intermediate layers, we propose to
extract internal components, i.e. feature facets, of self-attention blocks in
ViTs: queries, keys, and values.

\textbf{(2) Self-attention exploitation.} The architectural design of
self-attention empowers ViTs to capture semantic context of the image at a high
level. We propose, atop the adversarial exploitation of internal facets in ViT
blocks, to systematically extract saliency maps from the self-attention
mechanism itself, and integrate them into loss optimization as dense
semantic guides to identify valuable feature targets.

\textbf{(3) Joint self-supervision feature discrimination.} Two branches of
self-supervision paradigms exist for ViTs: contrastive learning (CL) and masked
image modeling (MIM). Comparative studies show that CL captures global
structural shapes and semantics, while MIM focuses more on local textural
details~\cite{ParkKH0Y23}. We hypothesis that, if combined, both aspects will
complement each other in generalizability that jointly contribute to enhancing
adversarial transferability.


Incorporating all three aspects, we introduce \dsvab{}---a generative
\underline{d}ual \underline{s}elf-supervised \underline{V}iT features
\underline{a}ttack. We introduce a novel generative training framework,
consisting of a generator to craft transferable adversarial perturbation, and
discriminative training approaches to jointly exploit the dual intricate
features---both structural and textural---distilled by the two types of
self-supervised ViTs. We choose the duo: DINO~\cite{CaronTMJMBJ21} and
MAE~\cite{HeCXLDG22}, for CL and MIM respectively. \Cref{fig:fig1}
showcases a birds-eye view of \dsva{}.

Leveraging the powerful latent representations distilled by self-supervised
ViTs, \dsva{} achieves outstanding adversarial effectiveness. We show an example
in \cref{fig:fig1} of \dsva{} successfully disrupting both structural features
from DINO (CL) and textural representations from MAE (MIM) (visualized in the
attention maps), enabling impressive transferability towards black-box models of
distinct architectures. Our experiments demonstrate \dsva{}'s outstanding
transferability to models across ViTs, ConvNets, and MLPs alike, and its ability
to evade defenses, surpassing various state-of-the-arts.

To conclude, we summarize our contributions as follows.

\begin{itemize}[wide]
  \item We present \dsvab{}, a generative adversarial attack, that crafts
    highly transferable black-box adversarial examples by exploiting dual
    self-supervised ViT features.
  \item We first aim at, instead of attacking the direct output of
    intermediate layers, targeting the internal facets of the self-attention
    blocks in ViTs, namely, the queries, keys, and values, to take advantage
    of the Transformer architecture and extract generalizable and
    transferable features.
  \item We further introduce a method to exploit the self-attention mechanism
    itself by extracting saliency maps from the self-attention maps of ViTs,
    acting as guides for important feature targets, providing, in essence, a
    regularization scheme that enable boosted adversarial generalizability.
  \item We finally propose to jointly exploit the two self-supervised
    learning schemes---CL and MIM---to craft perturbation that attend to and
    disrupt both global structural shapes and local textural details from
    within the image.
\end{itemize}


%% file: sec/2_relatedwork.tex
\section{Related Work}
\label{sec:related-work}

\begin{figure*}[t]
  \centering
  \includegraphics[width=1.0\linewidth]{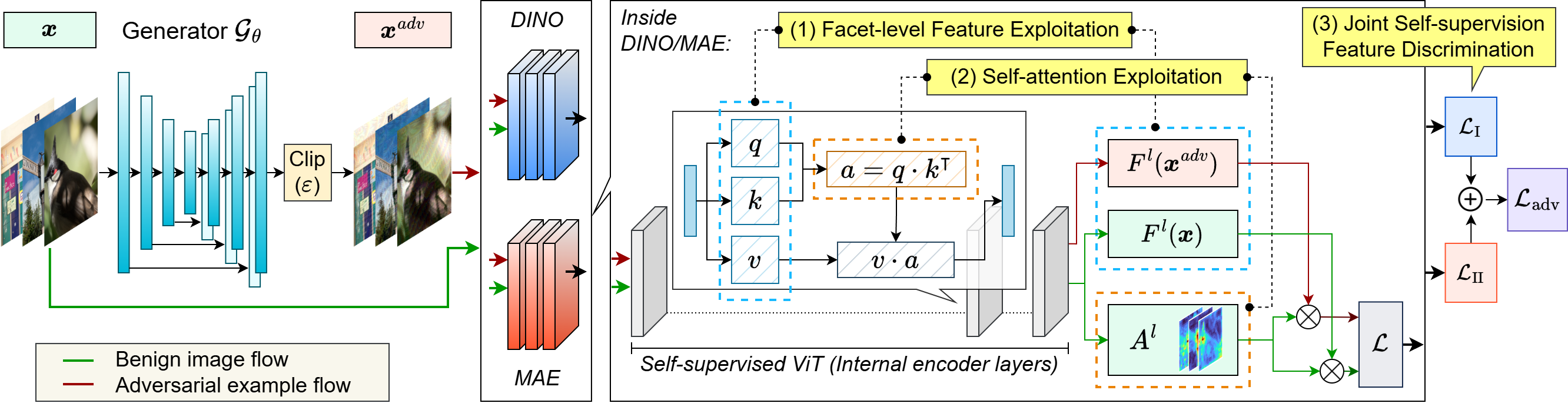}
  \caption{\textbf{The \dsva{} Training Framework.} Sample $\bx$ is fed
    through $\mathcal{G}$ to create adversarial example $\bxadv$, which are
    then both fed into the self-supervised models DINO and MAE, to extract
    deep representations and attention saliency maps from both global
    structural and local textural feature aspects. The feature
    discriminative loss is derived from both ViTs, which jointly form the
    adversarial loss $\mathcal{L}_\text{adv}$.
  }
  \label{fig:fig2}
\end{figure*}

\paragraph{Generative adversarial attacks} is initially introduced in
\citet{PoursaeedKGB18} to address both sample-agnostic and sample-specific
adversarial perturbation. This approach paved the way for generative methods in
creating unrestricted perturbations~\cite{SongSKE18} and utilizing
GANs~\cite{XiaoLZHLS18}. The generative strategy has further proven to be
beneficial for transferability, where \citet{Naseer0KKP19} developed CDA for
cross-domain attacks, \citet{NakkaS21} incorporated mid-level features, and
\citet{ZhangLCSGHX22} presented BIA for generating cross-domain perturbation
with only knowledge from ImageNet. We follow this foundational generative
approach in our work. Other studies refine the generator to improve
\textit{targeted} attack effectiveness~\cite{YangDPSZ22, FengXZ023, YFSGZR23} or
introduce \textit{outside knowledge} from foundation models trained on web-scale
datasets~\cite{YangJY24}. We do not consider them as our competitors.

\paragraph{Self-supervised learning} has enjoyed its remarkable success in
natural language processing, particularly with wide applications in modern
language models~\cite{DevlinCLT19, Radford2019LanguageMA}. In vision tasks,
although several self-supervised techniques have been developed for
ConvNets~\cite{CaronMMGBJ20, GrillSATRBDPGAP20, He0WXG20}, it is with ViTs that
the self-supervised learning strategy, through both CL~\cite{ChenK0H20,
ChenXH21, CaronTMJMBJ21, OquabDMVSKFHMEA24} and MIM~\cite{Xie00LBYD022,
Bao0PW22, HeCXLDG22, AssranDMBVRLB23}, has truly excelled. Self-supervised ViTs
have shown to encode rich features that carry incredible capabilities
out-of-the-box, often surpassing comparable methods that require additional
supervised finetuning~\cite{AmirGBD22, EngstlerMKRL2023, SeitzerHZZXS00S23}. In
this work, we propose to jointly exploit the dual aspects of features provided
in CL and MIM for crafting generalizable adversarial perturbation with superior
transferability. Note that we choose to use DINO~\cite{CaronTMJMBJ21} instead of
DINOv2~\cite{OquabDMVSKFHMEA24} for fair comparison, as DINOv2 is trained on a
far larger dataset than vanilla ImageNet.

%% file: sec/3_methodology.tex
\section{Methodology}
\label{sec:methodology}



\subsection{Threat Model}

We consider the standard $\ell_\infty$ threat model. Given a DNN classifier
$\mathcal{F}(\cdot): \bx \in \mathbb{R}^m \mapsto y$ where $\bx$ is a benign
sample and $y$ denotes its ground truth label. The adversary aims to create an
adversarial example $\bxadv = \bx + \delta$, with perturbation $\delta$
restricted by an $\ell_p$-ball ($\ell_\infty$ in our case), such that
$\mathcal{F}(\bxadv) \neq y$. We incorporate a generator $\mathcal{G}_{\btheta}$
to craft $\bxadv$ by discriminating the latent intermediate features of the
self-supervised ViTs as
\begin{equation}
  \btheta^* \leftarrow \underset{\btheta}{\arg\max}\ \mathcal{D}\left( F(\bx), F(\bxadv) \right),\ s.t.\ \lVert \delta \rVert_\infty \leq \varepsilon,
\end{equation}
where $\bxadv = \mathcal{G}_{\btheta}(\bx)$, $F(\cdot)$ extracts the
self-supervised ViT features from an image, and $\mathcal{D}(\cdot,\cdot)$
measures the feature distance. We now present our proposed \dsva{} for the
training of the adversarial generator $\mathcal{G}_{\btheta}$.

\subsection{Facet-level Feature Exploitation}

Previous arts have highlighted the strong transferability potential of
feature-space adversarial perturbation, but they focus on \textit{supervised
ConvNets}. In this work, we first explore the rich features offered by the
harmonic combination of self-supervision and the Transformer architecture.

Irrespective of training strategy, ViTs process images in the same manner. The
input image is divided into $n$ non-overlapping patches $\{ p_i \}\ (i\in
[1,n])$ and linearly projected onto a $D$-dimensional latent space. Positional
embeddings and the \texttt{[CLS]} token are added thereafter, forming a set of
tokens to be fed through $L$ layers of transformer encoders. Each encoder block
comprises alternating layers of multi-head self-attention (MSA) and MLP blocks,
with LayerNorm (LN) applied before each block. We denote the output token
sequence at layer $l$ as $T^l = \{ t_0^l, t_1^l, \cdots, t_n^l \}$.

If we were to follow previous practice, we would directly use intermediate
encoder layer outputs, i.e., tokens, as the feature representation. In contrast,
the Transformer architecture encodes features within MSA blocks that offer
better generalizability. At each layer $l$, the MSA block encodes tokens from
the previous layer $T ^ {l-1}$ into \textit{queries}, \textit{keys}, and
\textit{values}, i.e., $q_i^l = w_q^l \cdot t_i ^{l-1}$, $k_i^l = w_k^l \cdot
t_i ^{l-1}$, and $v_i^l = w_v^l \cdot t_i ^{l-1}$ (with $w ^ l$ being the
weights), which are fused back into $T ^ l$. Therefore, each image patch $p_i$
corresponds to a set of \textit{deep features} at the facet-level, namely $\{
q_i^l, k_i^l, v_i^l, t_i^l \}$, with each representing its internal query, key,
value, and the final output as a fused token at layer $l$. In ViTs, the
\textit{query} is the part of input the model is focusing on, whereas the
\textit{key} is then compared with the \textit{query} to determine the
attention. They are then aggregated into the \textit{value} vector for feature
concatenation. Facets \textit{key} and \textit{query} are directly associated
with the input, inherently providing high quality, less noisy features that
favor generalizability. We later empirically investigate the impact of facet
selection to adversarial effectiveness.



As in \cref{fig:fig2}, to train $\mathcal{G}_{\btheta}$ for perturbation
generation, \dsva{} is designed to deviate the latent representations of a
benign image and its generated adversarial example, that is, to minimize the
cosine similarity between the deep features extracted. In this way, the crafted
perturbation would be able to \textit{neutralize} critical decisive low-level
features within the sample, thereby misleading black-box DNNs. Hence, the
discriminative loss at this stage is formulated as
\begin{equation}\label{eq:2}
  \btheta^* \leftarrow \underset{\btheta}{\arg\min}\ \mathcal{D}_{\cos} \left(F^l \left(\bx \right), F^l \left(\bxadv \right) \right),
\end{equation}
where $F^l(\cdot)$ gives one of $q^l,k^l,v^l,t^l$ as the target facet-level
feature extracted at layer $l$ within the ViT $\mathcal{F}(\cdot)$. At inference
time, the trained generator $\mathcal{G}_{\btheta^*}$ crafts adversarial example
$\bxadv$ within perturbation budget as
\begin{equation}
  \bxadv = \mathrm{clip}(\mathcal{G}_{\btheta^*}(\bx), \varepsilon).
\end{equation}

\begin{figure*}[t]
  \centering
  \includegraphics[width=0.9\textwidth]{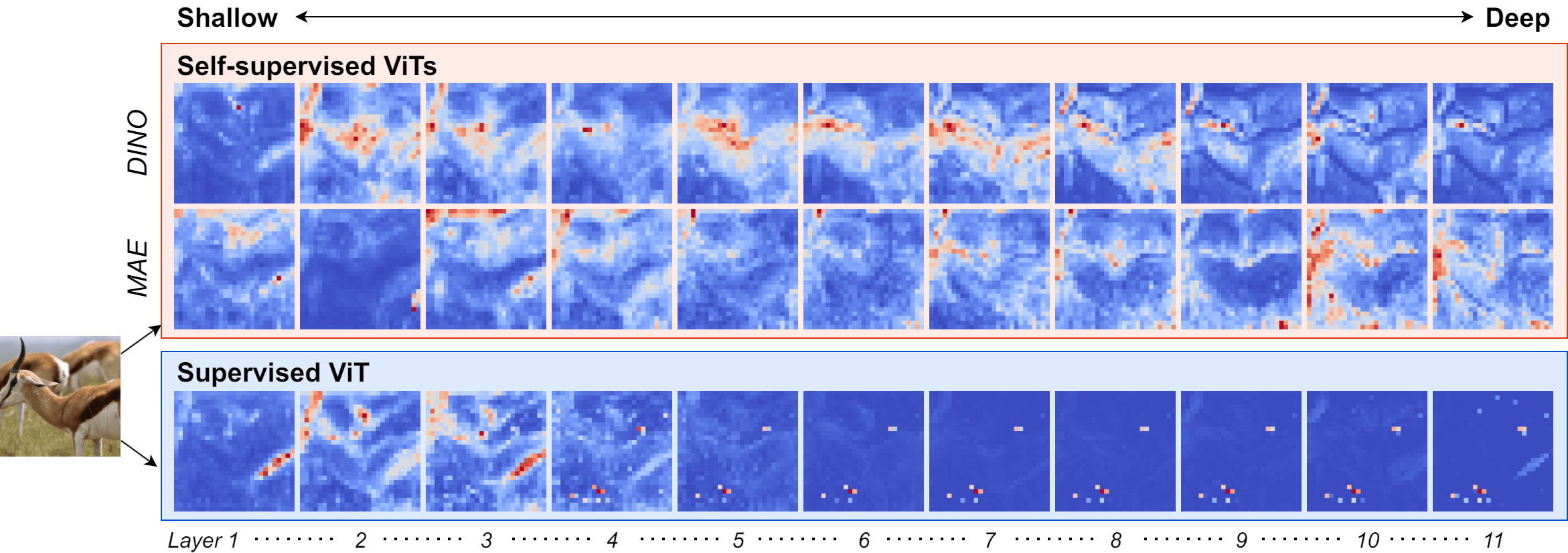}
  \caption{\textbf{Attention saliency maps.} We visualize the attention
    saliency maps derived from both self-supervised ViTs DINO (first row)
    and MAE (second row), and the supervised ViT (third row). From left to
    right, layer depth increase from shallow to deep (from 1 to 11).
  }
  \label{fig:fig3}
\end{figure*}

\subsection{Self-attention Exploitation}

\citet{CaronTMJMBJ21} revealed that the attention heads of self-supervised ViTs
attend to salient foreground regions in an image, and \citet{AmirGBD22} further
demonstrated that these encoded features represent powerful learnt common ground
across images. As such, we propose an incremental regularization to leverage
saliency maps derived from the self-attention mechanism of pretrained ViTs as
\textit{feature landmarks}, so as to offer additional guidance to target more
impactful features during optimization in \dsva{}.

We first extract the self-attention maps for benign sample $\bx$ at layer $l$,
i.e., the attention weights associated with each head of each token attending to
every other token, denoted as $A^l$. Next, we select the attention weights from
the \texttt{[CLS]} token to all other tokens across all heads as
\begin{equation}
  A^l_{\texttt{[CLS]}} = A^l [ :, :, 0, 1{:}].
\end{equation}
The attention saliency map $S^l$ at layer $l$ is calculated as the mean
attention from the \texttt{[CLS]} token to all other tokens over each attention
head as
\begin{equation}
  S^l = \frac{1}{H} \sum_{h=1}^H A^l_{\texttt{[CLS]}} [h],
\end{equation}
where $H$ is the number of attention heads. Thus, $S^l$ serves as a
\textit{feature landmark guidance} for targeting intermediate features,
regularizing the global semantic knowledge learnt by the generator. We apply a
scaling factor of $\gamma$ to $S^l$ for loss optimization. Building on
\cref{eq:2}, loss function $\mathcal{L}$ at this stage is thus formulated as
\begin{equation}\label{eq:6}
  \mathcal{L} = \underset{\btheta}{\arg\min}\ \mathcal{D}_{\cos} \left(F^l \left(\bx \right) \odot \left(\gamma \cdot S^l \right), F^l \left(\bxadv \right) \odot \left(\gamma \cdot S^l \right) \right).
\end{equation}

Shown in \cref{fig:fig3} is the attention saliency maps extracted from
ViTs with self-supervision (red background) vs. supervision (blue background),
as well as the variance of saliency maps with increasing layer depth from left
to right. Compared to the saliency maps extracted from a supervised ViT, those
from the self-supervised ViTs DINO (first row) and MAE (second row) are less
noisy and capture various levels of global and local semantics, respectively.
From shallow to deep layers, the self-supervised representations favor less
spatial information and more textural information, whereas the supervised ViT's
representations collapse into homogeneous primitive patterns. These
visualizations showcase the powerful representations offered only by the
self-attention of \textit{self-supervised ViTs}, acting as feature landmarks to
be integrated in \dsva{} for transferability boosts.

\subsection{Joint Self-supervision Feature Discrimination}

Recall that two branches of self-supervision strategies currently stand for
ViTs: CL and MIM. Reflected in both learnt latent representations and
self-attention favoritism, CL better captures global long-range shape-wise
features by learning globally projected representations to discriminate each
other, while MIM focuses more on local textural details as it is a generative
task that predicts masked regions. We hypothesis that features derived from CL
and MIM would complement each other from an adversarial perspective. Therefore,
we propose to jointly exploit both feature aspects in \dsva{} to disrupt
structure-biased and texture-biased image features, thereby enhancing
adversarial transferability.

\begin{table*}[t]
  \centering
  \resizebox{\textwidth}{!}{
    \begin{tabular}{*l^c^c^c^c^c^c^c^c^c^c^c^c}
      \toprule
      \rowstyle{\bfseries}
      Attack         & VGG-16  & Res-50  & Den-121 & Eff-B0  & Inc-v3  &
      Inc-v4  & Swin-B  & MaxViT  & PiT-B   & Visformer & LeViT   & Mixer
      \\
      \midrule
      CDA (VGG-19)   & \rb{99.31} & 69.23 & 59.19 & 76.38 & 52.94 & 61.96 & 16.53 & 14.63 & 9.48  & 32.40   & 29.79 & 23.02 \\
      CDA (Res-152)  & 92.98 & 88.88 & 87.02 & 75.32 & 63.85 & 74.97 & 11.82 & 7.78  & 5.86  & 39.03   & 35.85 & 22.78 \\
      CDA (Den-169)  & 92.98 & 87.63 & \rb{97.03} & 90.96 & 67.59 & 78.94 & 26.88 & 22.41 & 20.98 & 69.67   & 65.11 & 52.01 \\
      BIA (VGG-19)   & 97.58 & 74.32 & 84.93 & 77.77 & 66.63 & 76.96 & 19.35 & 15.25 & 12.46 & 34.68   & 35.96 & 27.53 \\
      BIA (Res-152)  & 94.94 & \rb{92.52} & 86.47 & 65.11 & 62.46 & 81.37 & 22.18 & 17.32 & 11.40 & 45.55   & 29.15 & 29.60 \\
      BIA (Den-169)  & 93.67 & 86.07 & 95.49 & 81.17 & 75.40 & 71.78 & 17.36 & 9.44  & 10.65 & 32.71   & 44.47 & 38.98 \\
      \midrule
      CDA (ViT-B/16) & 92.75 & 74.32 & 90.10 & 87.23 & 81.82 & 82.25 & \rb{62.13} & 33.09 & \rb{59.74} & 78.05   & 85.20 & 80.63 \\
      BIA (ViT-B/16) & 52.93 & 21.83 & 33.77 & 32.13 & 31.55 & 34.62 & 8.89  & 5.50  & 6.39  & 17.81   & 27.34 & 40.68 \\
      MI  (ViT-B/16) & 52.59 & 32.33 & 47.85 & 52.34 & 38.07 & 35.61 & 49.69 & 31.02 & 42.92 & 47.31   & 43.51 & 65.16 \\
      PNA (ViT-B/16) & 46.49 & 33.99 & 42.68 & 50.64 & 37.97 & 36.05 & 50.84 & 35.68 & 46.96 & 51.04   & 51.49 & 74.30 \\
      TGR (ViT-B/16) & 54.89 & 35.14 & 51.60 & 57.02 & 37.54 & 40.35 & 51.15 & 34.02 & 45.26 & 50.72   & 46.38 & 79.78 \\
      ATT (ViT-B/16) & 60.41 & 40.85 & 56.55 & 64.47 & 43.32 & 44.43 & 59.10 & 40.15 & 51.12 & 58.80 & 56.02 & 82.52 \\
      \midrule
      \dsva{} (DINO)   & 86.54 & 57.59 & 83.17 & 88.51 & 78.50 & 78.61 & 33.05 & 21.27 & 35.04 & 72.67 & 67.41 & 78.81 \\
      \dsva{} (MAE)    & 94.36 & 78.07 & 86.36 & 84.04 & 77.75 & 79.71 & 47.38 & 31.85 & 33.55 & 63.25 & 64.32 & 56.64 \\
      \rowcolor{gray!15}
      \dsva{} (Joint)  & 96.78 & 81.70 & 94.83 & \rb{95.32} &
      \rb{89.73} & \rb{91.73} & 59.83 & \rb{41.29} & 50.48 &
      \rb{81.37} & \rb{85.21} & \rb{85.38} \\
      \bottomrule
  \end{tabular}}
  \caption{\textbf{Comparison of black-box transferability.} We showcase the
    black-box fooling rate (\%) of \dsva{} and compared baseline attacks,
    against target black-box models with various architectures, including a
  total of 6 ConvNets, 5 ViTs, and an MLP-Mixer. }
  \label{tab:transfer-black-box}
\end{table*}

To this end, we jointly train generator $\mathcal{G}_{\btheta}$ against both CL
and MIM ViTs, i.e., DINO and MAE. The final loss function
$\mathcal{L}_{\text{adv}}$ is thus formulated as
\begin{equation}\label{eq:7}
  \mathcal{L}_{\text{adv}} = \lambda \cdot \mathcal{L}_\text{I} + (1 - \lambda)
  \cdot \mathcal{L}_\text{II},
\end{equation}
where $\mathcal{L}_\text{I}$ and $\mathcal{L}_\text{II}$ are derived as in
\cref{eq:6} from DINO and MAE, respectively. Doing so, \dsva{} is able to craft
highly transferable perturbation that targets both structural and textural image
features, greatly boosting transferability across various black-box models with
diverse architectures.

%% file: sec/4_experiments.tex
\section{Experiments}
\label{sec:experiments}

\subsection{Experimental Settings}

\paragraph{Datasets.} The training set of ImageNet with over 1.28 million samples
is used for training the generator. Following work that focus on
transferability, the dataset from \textit{NeurIPS 2017 Adversarial
Learning}~\cite{AlexyAADC2018}, comprising 1000 images from the ImageNet
validation set, is used for evaluation.

\paragraph{Implementation details.} ViT-B/16 architectures with default stride
$s=16$ is chosen for both the self-supervised DINO and MAE, and the normal
supervised variant. Pretrained weights on ImageNet are sourced from their
original implementations. Following baseline methods~\cite{Naseer0KKP19,
ZhangLCSGHX22}, we use the same ResNet generator for $\mathcal{G}_\theta$. It is
trained with the Adam optimizer with learning rate $\eta=2 {\times} 10^{-4}$
over a single epoch. Scaling factor of attention saliency map $\gamma=100$. We
report results of \dsva{} trained with (1) DINO only, (2) MAE only, and (3) both
DINO and MAE (Joint). \textit{(\dsva{} collapses to SVA when only one
    self-supervised ViT is used, but we stick to the name of \dsva{} to avoid
ambiguity.)}

\paragraph{Parameters.} For both DINO and MAE, we choose features extracted at
the penultimate layer $l=10$. We select the \textit{key} facet of DINO and the
\textit{query} facet of MAE to exploit. The joint training parameter of \dsva{}
is set as $\lambda=0.5$. The rationale and empirical evaluations supporting
these selections are presented in
\cref{sec:the-parameters,sec:visualizing-feature-disruption}.

\paragraph{Metric.} We employ the fooling rate, i.e., the ratio of the
adversarial examples which successfully fool the target model among all
generated samples, as the evaluation metric.

\paragraph{Attacks.} Generative attack baselines include
BIA~\cite{ZhangLCSGHX22} and CDA~\cite{Naseer0KKP19}. We use
VGG-19~\cite{SimonyanZ14a}, ResNet-152 (Res-152)~\cite{HeZRS16}, and
DenseNet-169 (Den-169)~\cite{HuangLMW17} as their surrogates with the same
perturbation budget of $\varepsilon=10$ to follow their setups. We also compare
against BIA and CDA trained on supervised ViT-B/16. We additionally include
evaluations against gradient-based attacks, including the classic MI-FGSM
(MI)~\cite{DongLPS0HL18}, and 3 other state-of-the-art attacks designed for ViTs
(PNA~\cite{WeiCGWGJ22}, TGR~\cite{ZhangHWL23}, ATT~\cite{MingRWF24}).
\textit{(In \cref{tab:transfer-black-box} and \cref{tab:transfer-defense},
    MI-FGSM is abbreviated as MI so as to avoid confusion with MIM---masked image
modeling.)}

\subsection{Transferability to Black-box Models}
\label{sec:transfer-black-box}

We first evaluate black-box transferability within the ImageNet domain. For
attack targets, we choose 3 ConvNets with the same structure as the surrogates
of the compared methods to follow baseline settings (VGG-16, ResNet-50 (Res-50),
DenseNet-121 (Den-121)). We add 3 ConvNets with a different structure
(EfficientNet-B0 (Eff-B0)~\cite{TanL19}, Inception-v3
(Inc-v3)~\cite{SzegedyVISW16}, Inception-v4 (Inc-v4)~\cite{SzegedyIVA17}), 5
ViTs (Swin-B~\cite{LiuL00W0LG21}, MaxViT-T~\cite{TuTZYMBL22},
  PiT-B~\cite{HeoYHCCO21}, VisFormer-S~\cite{ChenX00W021},
LeViT-128~\cite{GrahamETSJJD21}), and an MLP Mixer
(Mixer-B/16)~\cite{TolstikhinHKBZU21}. We report the results in
\cref{tab:transfer-black-box}.

Across all models, \dsva{} consistently achieves exceptional transferability,
outperforming baselines. As expected, BIA and CDA with surrogates VGG-19,
Res-152, and Den-169 slightly outperforms \dsva{} on VGG-16, Res-50, and
Den-121, as they share the same structure. Nevertheless, the transferability of
\dsva{} (Joint) surpasses all compared attacks on the remaining models,
particularly non-ConvNets. Even when using a \textit{supervised ViT surrogate},
competing attacks fail to match \dsva{}'s performance, including
state-of-the-art attacks that are tailored for ViTs. Only CDA with a supervised
ViT matches \dsva{} in 2 cases (Swin-B and PiT-B). Our results show that (1)
without our proposed exploitation schemes in \dsva, existing feature-level
attacks simply cannot take full advantage of the Transformer architecture, and
(2) \dsva{} (Joint) outperforms its single model variants by 13.70\% on average,
underscoring the importance of our joint exploit of the complementary structural
and textural features from the self-supervised strategy duo.

\subsection{Transferability to Defense Models}
\label{sec:transfer-defense}

\begin{table}[t]
  \centering
  \resizebox{\linewidth}{!}{
    \begin{tabular}{*l^b{0.78cm}^b{0.78cm}^b{0.78cm}^b{0.88cm}^b{0.88cm}^b{0.78cm}}
      \toprule
      \rowstyle{\bfseries}
      Attack         & Inc-v3\textsubscript{adv} &
      Inc-v3\textsubscript{ens3} & Inc-v4\textsubscript{ens4} &
      IncRes-v2\textsubscript{ens} & IncRes-v2\textsubscript{adv} &
      Eff-b0\textsubscript{ap} \\
      \midrule
      CDA (VGG-19)   & 25.05    & 16.36     & 9.78      & 10.73       & 34.90       & 67.39   \\
      CDA (Res-152)  & 43.01    & 38.60     & 28.88     & 29.27       & 61.89       & 73.91   \\
      CDA (Den-169)  & 53.44    & 41.11     & 27.08     & 24.58       & 66.00       & 83.33   \\
      BIA (VGG-19)   & 39.57    & 28.35     & 21.24     & 17.60       & 62.19       & 79.71   \\
      BIA (Res-152)  & 32.26    & 27.15     & 19.89     & 17.50       & 63.29       & 70.29   \\
      BIA (Den-169)  & 55.91    & 43.40     & 37.64     & 30.52       & 59.08       & 86.23   \\
      \midrule
      CDA (ViT-B/16) & 65.91    & 53.98     & 50.67     & 38.54       & 71.11       & 86.23   \\
      BIA (ViT-B/16) & 22.80    & 15.38     & 12.02     & 10.83       & 24.97       & 52.17   \\
      MI (ViT-B/16)  & 26.67    & 22.46     & 21.91     & 18.85       & 26.98       & 55.07   \\
      PNA (ViT-B/16) & 27.63    & 22.90     & 22.70     & 19.79       & 29.69       & 55.07   \\
      TGR (ViT-B/16) & 30.22    & 25.85     & 24.83     & 21.67       & 29.89       & 67.39   \\
      ATT (ViT-B/16) & 40.43    & 36.21     & 33.03     & 29.79       & 41.52       & 75.36 \\
      \midrule
      \dsva{} (DINO)   & 66.13    & 54.09     & 49.33     & 43.85       &
      75.03       & \rb{89.96}   \\
      \dsva{} (MAE)    & 50.11    & 32.39    & 28.88     & 23.85       & 66.70       & 76.09   \\
      \rowcolor{gray!15}
      \dsva{} (Joint)  & \rb{79.03}    & \rb{68.16}     &
      \rb{62.70}     & \rb{52.50}       & \rb{88.06}       &
      89.13   \\
      \bottomrule
  \end{tabular}}
  \caption{\textbf{Comparison of transferability against models with defenses.}
    We report the black-box fooling rate (\%) of \dsva{} and compared
    baseline attacks in defenses evasion, on various models with adversarial
  training enabled within ImageNet.}
  \label{tab:transfer-defense}
\end{table}

Next, we validate our approach against defenses, an aspect previously unexplored
in the context of generative attacks. We follow previous
setups~\cite{WeiCGWGJ22, ZhangHWL23, MingRWF24} and use 6 robust black-box
models on ImageNet to evaluate defense evasion, namely
Inc-v3\textsubscript{adv}, IncRes-v2\textsubscript{adv}~\cite{KurakinGB17},
Inc-v3\textsubscript{ens3}, Inc-v4\textsubscript{ens4},
IncRes-v2\textsubscript{ens}~\cite{TramerKPGBM18}, and EfficientNet-B0 with
AdvProp~\cite{XieTGWYL20} (Eff-b0\textsubscript{ap}). Shown in
\cref{tab:transfer-defense}, we once again observe that \dsva{} shows superior
performance across all adversarially trained models, with \dsva{} (Joint)
achieving transferability that exceeds all compared attacks by an average margin
of 32.98\%. We contend that while adversarial training enhances DNN robustness
by developing more resilient features, they ultimately need to use these same
essential features for classification. By fully exploiting self-supervised ViT
representations, decisive elements of the sample are destroyed at a more
generalized level, allowing \dsva{} to evade these defenses. We provide
additional results against state-of-the-art defenses and robust ViTs in Appendix
C.

\subsection{Analysis on the Impact of Relevant Parameters}
\label{sec:the-parameters}

\begin{figure}[t]
  \centering
  \includegraphics[width=\linewidth]{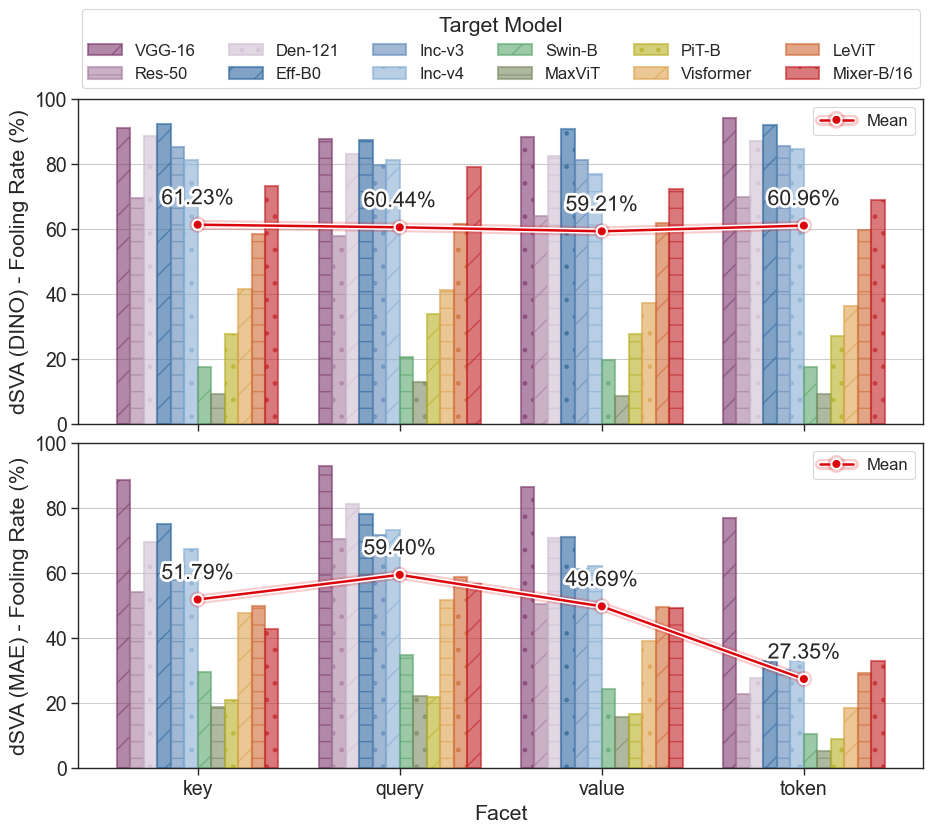}
  \caption{\textbf{Impact of the choice of facet.} We evaluate the
    transferability of \dsva{} (DINO) and \dsva{} (MAE) that exploit
    feature facets at layer 10 of \textit{query}, \textit{key},
  \textit{value}, and \textit{token}, respectively.}
  \label{fig:variable-facet}
\end{figure}

\begin{figure}[t]
  \centering
  \includegraphics[width=\linewidth]{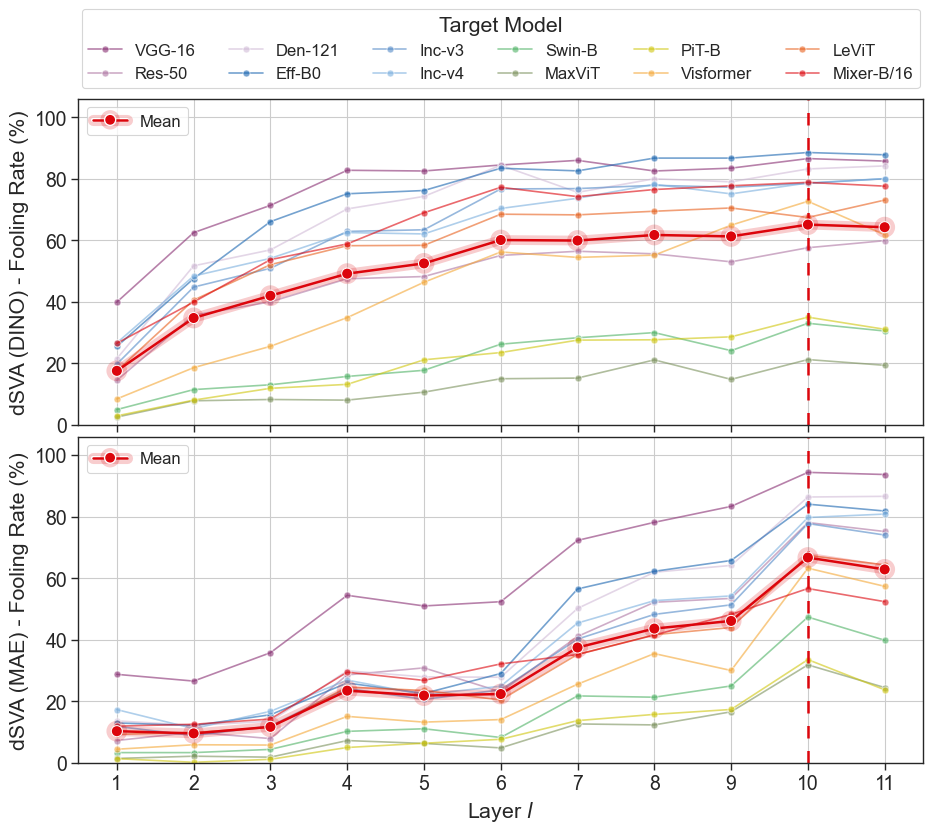}
  \caption{\textbf{Impact of the choice of layer $l$.} We evaluate the
    transferability of \dsva{} (DINO) and \dsva{} (MAE) with layer $l$ from
  1 to 11 (from left to right).}
  \label{fig:variable-layer}
\end{figure}

\begin{figure}[t]
  \centering
  \includegraphics[width=\linewidth]{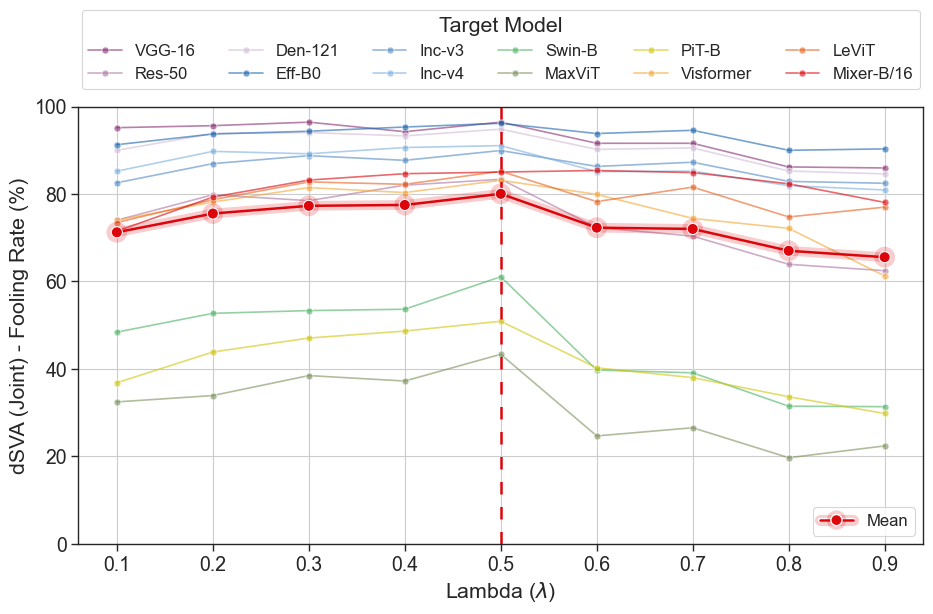}
  \caption{\textbf{Impact of the choice of $\lambda$.} We evaluate the
    transferability of \dsva{} (Joint) with default parameters employed
    except for $\lambda$. $\lambda$ is applied from 0 to 1 with a step size
  of 0.1.}
  \label{fig:variable-lambda}
\end{figure}

We now turn our focus to \dsva{}'s deciding \textit{parameters}, that is, (1)
feature facet (\textit{query}, \textit{key}, \textit{value}, or the entire
layer's output: \textit{token}), (2) feature layer $l$, and (3) $\lambda$, for
\dsva{} (Joint). \textit{(In
    \cref{fig:variable-facet,fig:variable-layer,fig:variable-lambda} of
    \cref{sec:the-parameters}, the bold red line represents the mean transferability
    of the evaluated variant of \dsva{}, aggregated over observations against target
models.)}

\paragraph{The choice of facet $\{ q, k, v, t \}$.} We first evaluate the
performances of \dsva{} (DINO) and \dsva{} (MAE) with respect to the exploited
facets. We report the black-box transferability of them in
\cref{fig:variable-facet}. We first note that the variants that directly exploit
the \textit{token} facet, i.e, the entire intermediate layer output, always lags
behind, especially in the case of \dsva{} (MAE). These findings underline the
efficacy of our proposed facet-level exploit to capitalize on the adversarial
potential of the features distilled by the Transformer architecture. For MAE,
the \textit{query} directly serves as the input with masked patches, which is
intuitively more crucial for its reconstruction task. The \textit{key} facet in
this case only provides additional \textit{context} of the current masked
modelling session. This aligns with our observation that \dsva{} (MAE) performs
best with the \textit{query} facet. For DINO, the student network generates one
view of the image as the \textit{query}, while the teacher uses another as the
\textit{key}. The teacher, acting as a guide, would provide a better contrastive
signal. Our results, although not as pronounced as the MAE variant, show that
\dsva{} (DINO) performs best when exploiting the \textit{key} facet.

\paragraph{The choice of layer $l$.} Next, we investigate the impact of layer
$l$. We report \dsva{} (DINO) and \dsva{} (MAE)'s transferability that exploit
layer $l$ from 1 to 11 in \cref{fig:variable-layer}. We notice that the
transferability of \dsva{} tends to increase as layer deepens. We reason that as
the layers deepen, both self-supervised strategies manage to encode richer and
more generalizable semantic information, benefiting adversarial transferability.
Notably, transferability of both variants drops at the final 11th layer. This is
expected as the final layer of Transformer-based models is often optimized for
specific training setups, which results in significant reduction in
generalizability~\cite{DevlinCLT19}. In terms of vision tasks, ViTs have also
shown to maintain spatial and positional information in all but the last
layer~\cite{JelassiSL22, GhiasiKB2212}. We choose the penultimate layer of
$l=10$ of both DINO and MAE in \dsva{}.

\begin{figure*}[t]
  \centering
  \includegraphics[width=0.92\linewidth]{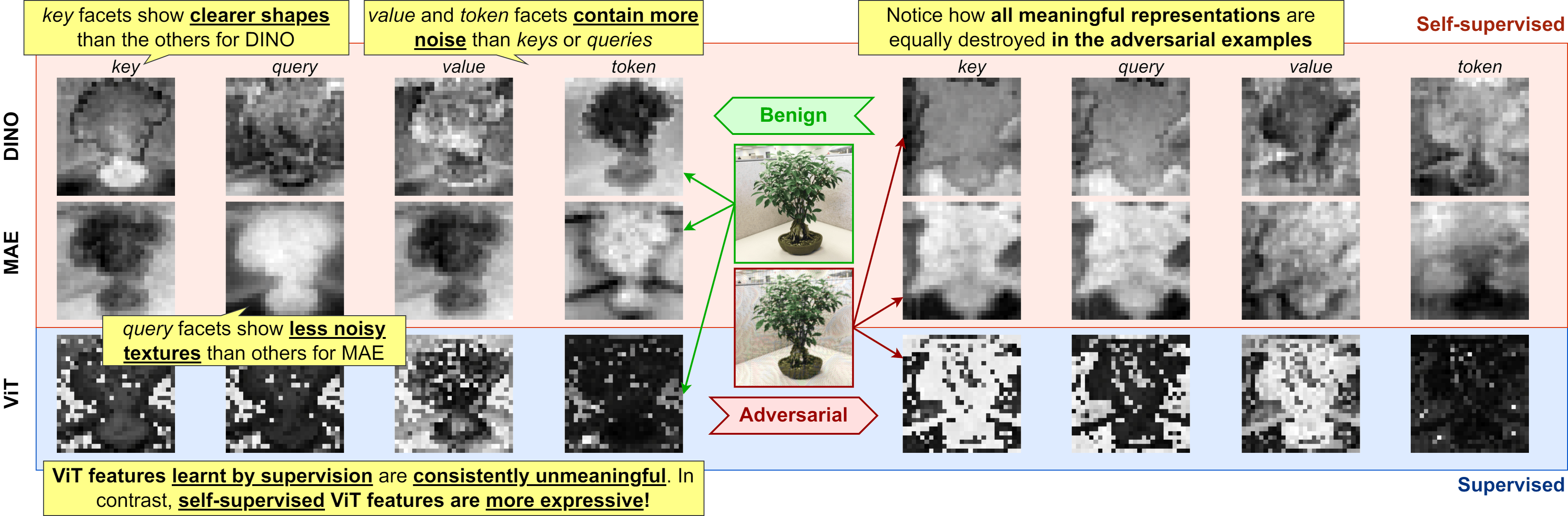}
  \caption{\textbf{Visualization of feature disruption.} We present PCA
    visualizations of the features extracted from all facets of DINO, MAE,
    and supervised ViT-B/16. Features of benign images are shown on the
  left, and adversarial examples crafted by \dsva{} (Joint) on the right.}
  \label{fig:pca}
\end{figure*}

\begin{figure*}[t]
  \centering
  \includegraphics[width=0.93\linewidth]{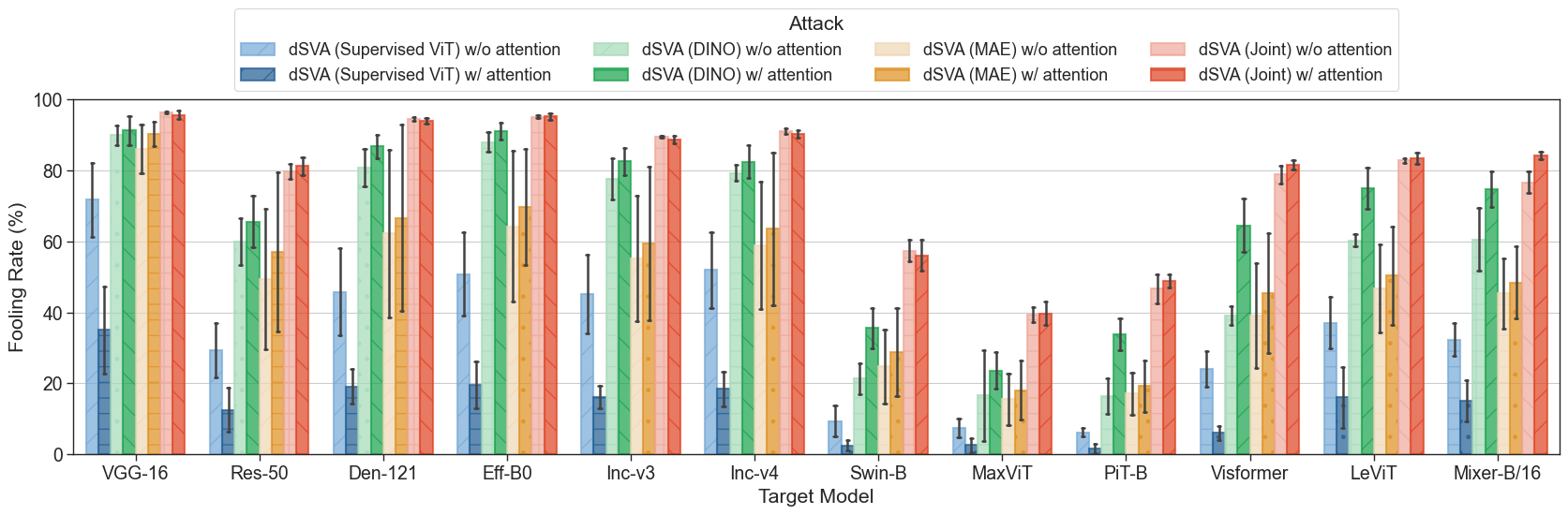}
  \caption{\textbf{Ablation study.} We present comparisons of the
    transferability of \dsva{} with supervised ViT, DINO, MAE, and Joint
    variants, with and without our proposed attention regularization
    applied, respectively. Results are aggregated over multiple
  observations.}
  \label{fig:ablation}
\end{figure*}

\paragraph{The choice of joint training parameter $\lambda$.} We finally explore
the key factor of \dsva{} (Joint), that is, the balance between feature
disruption for DINO (CL) and MAE (MIM), which is controlled by $\lambda$ as
described in \cref{eq:7}. The transferability of \dsva{} (Joint) with $\lambda$
in $(0, 1)$ with a step size of 0.1 is reported in \cref{fig:variable-lambda}.
We observe two interesting trends. First, as the dual aspects of features are
more incorporated into \dsva{} (as $\lambda$ approaches the midpoint),
adversarial effectiveness increases. This behavior substantiates our hypothesis
that the features provided by CL and MIM complement each other under an
adversarial context, where both global and local relationships are to be
destroyed, highlighting the importance of our proposed joint feature disruption.
In addition, as $\lambda$ decreases from 0.9 to 0.5, that is, as the aspect of
MIM features increase while CL features decrease, adversarial effectiveness show
a tendency to rise. We argue that the while CL provided structures are crucial
for shape/object distinction from a human standpoint, to craft generalized
perturbation for fooling DNNs, textural details distilled by MIM ought to be
more purposefully considered, as DNNs favor these fine-grained details.
$\lambda=0.5$ yields the best performances in our setup, but given the
similarity of the trends for $\lambda$ in $[0.3, 0.5]$, we suggest that the
optimal $\lambda$ may vary depending on the specific task or dataset.

\subsection{Visualizing Facet-level Feature Disruption}
\label{sec:visualizing-feature-disruption}

In \cref{fig:pca}, we visualize how self-supervised ViT features are more
meaningful than supervised ones, and how some ViT feature facets are more
crucial than others. We conduct PCA on DINO, MAE, and supervised ViT-B/16's
features on all facets. We notice once again that the self-supervised features
are richer and less noisy than the supervised ones. We find that, for both DINO
and MAE, the \textit{value} and \textit{token} facets are noisier than the
\textit{query} and \textit{key} facets. For DINO, its \textit{key} facet shows
more distinct shapes and objects, whereas for MAE, its \textit{query} facet
shows less noisy textured details. These observations align with our parameter
selections. We also show how \dsva{}'s adversarial perturbation equally destroys
meaningful semantics within the image, underscoring our approach's effectiveness
in feature disruption.

\subsection{Ablation Study}
\label{sec:ablation}

We finally conduct an ablation study on two factors: (1) self-supervision, and
(2) self-attention exploitation. We report the transferability of \dsva{} with
supervised ViT, DINO, MAE, and Joint variants, both w/ and w/o attention
saliency map regularization applied, in \cref{fig:ablation}. For single
model variants, we aggregate the results over all facets. For \dsva{} (Joint),
we aggregate the observations over $\lambda \in [0.3, 0.5]$.

\paragraph{Self-supervision.} When comparing \dsva{} variants w/ self-supervised
features to the supervised variant under identical conditions, even single model
variants, \dsva{} (DINO) and \dsva{} (MAE), outperform the supervised version
across all models. We once again showcase that the synergy between
self-supervision and the Transformer architecture, the central motivation of our
work, pushes adversarial effectiveness to a new level, heightening the
capability of our approach.

\paragraph{Self-attention exploitation.} We first observe that the
self-attention of supervised ViTs actually impair adversarial effectiveness when
applied as a regularization. As previously shown, attention saliency maps
extracted from the supervised ViT fail to match its self-supervised counterparts
for feature landmark guidance. \dsva{} with self-supervised ViTs DINO and MAE
consistently perform better when self-attention is also exploited. While \dsva{}
(Joint) outperforms all single model variants, its transferability occasionally
slightly degrades when attention regularization is applied, particularly when
transferability is already high. We find that \dsva{} (Joint) works best with
attention regularization active when targeting stronger or more sophisticated
models.

\subsection{Cross-domain Transferability}
\label{sec:cross-domain-transfer}

Our major competitors BIA and CDA show strong cross-domain transferability with
only ImageNet domain knowledge. We provide additional comparisons under
cross-domain settings in Appendix B. Results show that \dsva{} still maintains
superior transferability to both coarse and fine-grained classification domains
in most cases, offering boosts of approximately 6\% on average.

%% file: sec/5_conclusion.tex
\section{Conclusion}
\label{sec:conclusion}

We present a novel generative adversarial attack, \dsva{}, that successfully
exploits deep intermediate features distilled through the self-supervised
learning of ViTs. By aiming at facet-level feature representations, \dsva{}
takes full advantage of the ViT's internal architecture. With self-attention
regularization, \dsva{} vigilantly focuses on salient feature targets that are
valuable for exploitation. Through our joint disruption of both structural and
textural representations distilled by the self-supervised learning duo---CL and
MIM---\dsva{} crafts remarkably generalizable perturbation, achieving
state-of-the-art transferability. We demonstrate, through extensive experiments,
the superior adversarial transferability of \dsva{} to various black-box DNNs of
distinct architectures. \textit{Our research strongly indicates that effective
  adversarial exploitation of ViTs, especially feature-wise, is very much muted by
the use of surrogate models constrained by supervised learning.} We believe this
work encourages further exploration of the robustness implications of DNNs
within a self-supervised learning context.

\paragraph{Ackowledgements.} This work was supported by the National Natural
Science Foundation of China (U2336201).

%% file: sec/6_suppl.tex
\clearpage
\appendix

\section{Experimental Details}
\label{sec:experimental-details}

In this section, we disclose the details of our experimental evaluations
regarding the specific computational resources utilized, including hardware,
memory, and time consumption. All our experimental evaluations are all conducted
on GPU compute units equipped with an 11th Gen Intel(R) Core(TM) i9-11900K CPU,
a single NVIDIA GeForce RTX 4090 GPU, and 128 GB of onboard memory.

For \dsva{} with DINO, MAE, and the vanilla supervised ViT-B/16 at a stride of
$s=16$, as well as for all compared generative attacks (CDA, BIA), generator
$\mathcal{G}_\theta$ is trained on the entirely of the ImageNet training set for
one epoch with a batch size of 32. Under this setup, single model variants of
\dsva{} require up to 4 hours of training, \textit{a duration comparable to
previous methods.} For the joint variant, i.e., \dsva{} (Joint), batch size is
set to 22, where its training takes up to 7 hours to complete. Our proposed
additional exploit of self-attention (which is optional) in \dsva{} does not
increase the training time. The inference time for the adversarial generator is
comparable to, if not faster than, that of gradient-based iterative adversarial
attacks. For all settings, GPU memory utilization approximates to over 90\%. We
organize the rest of the experimental details in
\cref{tab:experimental-details}, which includes ViTs with stride of $s=8$ that
we use in sections that report results of cross-domain transferability.

\begin{table}[h]
  \centering
  \resizebox{\linewidth}{!}{
    \begin{tabular}{*l^b{1.2cm}^b{1.2cm}^b{1.6cm}^b{1.6cm}}
      \toprule
      \rowstyle{\bfseries}
      Attack & Stride $s$ & Batch Size & GPU Memory & Training Time \\
      \midrule
      \dsva{} (DINO)  & 16 & 32 & > 90\% & \textasciitilde  4 hours \\
      \dsva{} (DINO)  & 8  & 12 & > 90\% & \textasciitilde 13 hours \\
      \dsva{} (MAE)   & 16 & 32 & > 90\% & \textasciitilde  4 hours \\
      \dsva{} (MAE)   & 8  & 12 & > 90\% & \textasciitilde 13 hours \\
      \dsva{} (Joint) & 16 & 22 & > 90\% & \textasciitilde  7 hours \\
      \dsva{} (Joint) & 8  & 6  & > 90\% & \textasciitilde 25 hours \\
      \bottomrule
  \end{tabular}}
  \caption{\textbf{Computational resource details of our experiments.} We
    report the computational details of all variants of \dsva{} with
  different ViT configurations that we evaluate.}
  \label{tab:experimental-details}
\end{table}

\section{Results of Cross-domain Transferability}
\label{sec:results-cross-domain-transferability}

In this section, we provide supplemental experimental results on the
cross-domain transferability of \dsva{} in both coarse and fine-grained
classification tasks. The evaluations follow the baseline settings specified in
previous work~\cite{ZhangLCSGHX22}. For coarse-grained classification, we
evaluate both attacks on target black-box domains, namely, CIFAR-10,
CIFAR-100~\cite{Krizhevsky2009LearningML}, SVHN~\cite{Netzer2011ReadingDI}, and
STL-10~\cite{Coates2011AnAO}, with the same models. For fine-grained
classification, we report black-box transferability across three fine-grained
domains: CUB-200-2011~\cite{WahCUB2002011}, Stanford Cars~\cite{Krause0DF13},
and FGVC Aircraft~\cite{maji13fine-grained}. For each domain, we evaluate
against three black-box ConvNets with ResNet-50 (Res-50), SENet154, and
SE-ResNet101 (SE-Res-101) backbones, trained using the DCL
framework~\cite{ChenBZM19}.

\begin{table}[t]
  \centering
  \resizebox{\linewidth}{!}{
    \begin{tabular}{*l^r^l^p{1cm}^p{1cm}^p{1cm}^p{1cm}}
      \toprule
      \rowstyle{\bfseries}
      \multirow{2}{*}[-10pt]{Attack}
      & \multirow{2}{*}[-10pt]{$s$}
      & \multirow{2}{*}[-10pt]{$A$}
      & \multicolumn{4}{c}{\bfseries Domain} \\
      \cmidrule{4-7} & & & CIFAR-10 & CIFAR-100 & SVHN & STL-10 \\
      \midrule
      CDA (VGG-19)  & / & / & 12.65 & 30.79 & 3.36 & 7.56 \\
      CDA (Res-152) & / & / & 10.34 & 28.23 & 5.49 & 6.15 \\
      CDA (Den-169) & / & / & 27.42 & 53.22 & 6.84 & 10.31 \\
      BIA (VGG-19)  & / & / & \rb{39.04} & \rb{68.25} & 6.38  & 9.84  \\
      BIA (Res-152) & / & / & 26.24 & 49.36 & 3.75  & 7.35  \\
      BIA (Den-169) & / & / & 22.05 & 45.82 & \rb{12.79} & \rb{10.75} \\
      \midrule
      \dsva{} (DINO) & 16 & w/o & 13.98 & 37.67 & \rb{12.88} & 11.07 \\
      \dsva{} (DINO) & 8  & w/o & 24.05 & 53.00 & 6.54  & 11.18 \\
      \dsva{} (DINO) & 16 & w/ & 13.34 & 37.42 & 9.30  & \rb{12.66} \\
      \dsva{} (DINO) & 8  & w/ & 21.94 & 48.94 & 7.53  & 10.70 \\
      \dsva{} (MAE)  & 16 & w/o & 16.89 & 35.80 & 6.80  & 10.41 \\
      \dsva{} (MAE)  & 8  & w/o & 24.77 & 41.15 & 9.13  & 10.26 \\
      \dsva{} (MAE)  & 16 & w/ & 17.47 & 34.32 & 4.91  & 9.31  \\
      \dsva{} (MAE)  & 8  & w/ & 24.30 & 44.61 & 6.74  & 11.44 \\
      \rowcolor{gray!15}
      \dsva{} (Joint) & 16 & w/o & 23.64 & 50.28 & 8.94  & 11.04 \\
      \rowcolor{gray!15}
      \dsva{} (Joint) & 8  & w/o & \rb{26.87} & \rb{55.53} & 8.83  & 12.42 \\
      \rowcolor{gray!15}
      \dsva{} (Joint) & 16 & w/ & 21.56 & 43.25 & 8.82  & 11.89 \\
      \rowcolor{gray!15}
      \dsva{} (Joint) & 8  & w/ & 24.13 & 46.73 & 11.73 & 11.95 \\
      \bottomrule
  \end{tabular}}
  \caption{\textbf{Transferability towards coarse-grained classification
    domains.} We report transferability (\%) towards domains CIFAR-10,
    CIFAR-100, SVHN, and STL-10. $s$ is the stride of ViT-B/16. $A$ denotes
  whether attention regularization in \dsva{} is activated.}
  \label{tab:transfer-cross-domain-coarse-grained}
\end{table}

\begin{table*}[t]
  \centering
  \resizebox{\linewidth}{!}{
    \begin{tabular}{lllccccccccc}
      \toprule
      \rowstyle{\bfseries}
      \multirow{2}{*}[-3pt]{Attack}
      & \multirow{2}{*}[-3pt]{$s$}
      & \multirow{2}{*}[-3pt]{$A$}
      & \multicolumn{3}{c}{\bfseries CUB-200-2011}
      & \multicolumn{3}{c}{\bfseries Stanford Cars}
      & \multicolumn{3}{c}{\bfseries FGVC Aircraft} \\
      \cmidrule{4-12}
      & & & Res-50 & SENet154 & SE-Res-101 & Res-50 & SENet154 & SE-Res-101 & Res-50 & SENet154 & SE-Res-101\\
      \midrule
      CDA (VGG-19)  & / & / & 29.49 & 29.94 & 20.79 & 21.84 & 20.95 & 10.42 & 24.81 & 40.91 & 23.02 \\
      CDA (Res-152) & / & / & 49.85 & 48.77 & 34.77 & 48.08 & 37.91 & 21.60 & 33.80 & 48.01 & 36.19 \\
      CDA (Den-169) & / & / & 39.55 & 29.52 & 36.40 & 42.16 & 25.26 & 19.22 & 30.61 & 32.92 & 33.77 \\
      BIA (VGG-19)  & / & / & 62.21 & 52.78 & 36.84 & 70.93 & 37.01 & 29.86 & 82.61 & 51.17 & 51.27 \\
      BIA (Res-152) & / & / & 63.53 & 68.15 & 38.92 & 56.91 & 58.49 & 19.03 & 41.52 & 77.61 & 42.33 \\
      BIA (Den-169) & / & / & \rb{83.36} & 65.75 & 45.77 & \rb{91.67} & 51.75 & 52.57 & \rb{96.16} & 59.78 & 65.22 \\
      \midrule
      \dsva{} (DINO)  & 16 & w/o & 38.86 & 51.65 & 43.66 & \rb{53.57} & 59.22 & 50.79 & \rb{72.52} & 81.45 & 64.73 \\
      \dsva{} (DINO)  & 8  & w/o & 71.18 & 61.15 & 59.57 & 49.39 & 59.76 & \rb{56.23} & 54.38 & 77.71 & 67.96 \\
      \dsva{} (DINO)  & 16 & w/ & 41.55 & 49.48 & 47.75 & 47.01 & 51.25 & 47.23 & 53.57 & 61.83 & 66.10 \\
      \dsva{} (DINO)  & 8  & w/ & 33.68 & 40.99 & 38.12 & 33.78 & 37.92 & 29.92 & 37.12 & 46.25 & 55.68 \\
      \dsva{} (MAE)   & 16 & w/o & 42.93 & 51.81 & 37.56 & 28.80  & 47.10 & 20.24 & 34.13 & 50.62 & 43.86 \\
      \dsva{} (MAE)   & 8  & w/o & 37.38 & 58.97 & 36.44 & 44.28 & 38.30 & 26.74 & 29.70  & 50.10 & 36.58 \\
      \dsva{} (MAE)   & 16 & w/ & 60.08 & 63.80 & 42.42 & 41.22 & 62.48 & 26.79 & 38.81 & 72.95 & 57.45 \\
      \dsva{} (MAE)   & 8  & w/ & 42.38 & 62.11 & 41.99 & 46.04 & 38.99 & 29.33 & 30.41 & 52.90 & 43.73 \\
      \rowcolor{gray!15}
      \dsva{} (Joint) & 16 & w/o & \rb{78.77} & 79.62 & 66.11 & 48.67 & \rb{68.47} & 51.97 & 65.65 & 89.24 & \rb{83.15} \\
      \rowcolor{gray!15}
      \dsva{} (Joint) & 8  & w/o & 62.58 & 72.17 & 59.11 & 41.42 & 55.68 & 41.17 & 46.76 & 75.07 & 63.62 \\
      \rowcolor{gray!15}
      \dsva{} (Joint) & 16 & w/ & 76.44 & \rb{79.64} & \rb{69.72} & 47.29 & 67.91 & 50.99 & 68.94 & \rb{89.93} & 77.37 \\
      \rowcolor{gray!15}
      \dsva{} (Joint) & 8  & w/ & 70.88 & 78.85 & 68.24 & 47.25 & 66.30 & 50.12 & 68.15 & 87.97 & 74.10 \\
      \bottomrule
  \end{tabular}}
  \caption{\textbf{Transferability towards fine-grained classification
    domains.} We report transferability (\%) towards domains CUB-200-2011,
    Stanford Cars, and FGVC Aircraft. $s$ is the stride of ViT-B/16. $A$
  denotes whether attention regularization in \dsva{} is activated.}
  \label{tab:transfer-cross-domain-fine-grained}
\end{table*}

\begin{table*}[htbp]
  \centering
  \resizebox{\linewidth}{!}{
    \begin{tabular}{l>{\raggedleft\arraybackslash}b{1.8cm}>{\raggedleft\arraybackslash}b{1.8cm}>{\raggedleft\arraybackslash}b{1.8cm}>{\raggedleft\arraybackslash}b{1.8cm}>{\raggedleft\arraybackslash}b{1.8cm}>{\raggedleft\arraybackslash}b{1.8cm}>{\raggedleft\arraybackslash}b{2.0cm}>{\raggedleft\arraybackslash}b{2.0cm}}
      \toprule
      \textbf{Attack}
      & \textbf{Res-18} \cite{salman2020adversarially}
      & \textbf{Res-50} \cite{wongfast}
      & \textbf{ViT-B} \cite{mo2022adversarial} & \textbf{Swin-B} \cite{mo2022adversarial}
      & \textbf{XCiT-S12} \cite{debenedetti2023light}
      & \textbf{ViT-S +ConvStem} \cite{singh2023revisiting}
      & \textbf{ConvNeXt +ConvStem} \cite{singh2023revisiting}
      & \textbf{ConvNeXt-v2+Swin-L} \cite{bai2024mixednuts} \\
      \midrule
      CDA (VGG-19) & 7.13  & 8.25  & 6.09  & 10.15 & 7.91  & 6.69  & 4.96  & 5.68  \\
      CDA (Res-152) & 12.56 & 11.39 & 12.31 & 13.20 & 10.74 & 7.39  & 7.04  & 7.07  \\
      CDA (Den-169) & 11.21 & 12.54 & 9.96  & 16.38 & 13.93 & 10.33 & 8.19  & 8.89  \\
      BIA (VGG-19) & 12.05 & 11.22 & 8.85  & 12.96 & 11.22 & 9.51  & 7.50  & 7.50  \\
      BIA (Res-152) & 16.13 & 15.35 & 14.52 & 19.32 & 16.06 & 11.97 & 10.61 & 8.24  \\
      BIA (Den-169) & 14.09 & 14.19 & 18.95 & 22.62 & 16.65 & 10.92 & 9.80  & 9.42  \\
      \midrule
      CDA (ViT-B/16) & 12.39 & 13.04 & 8.85  & 18.70 & 14.52 & 11.39 & 9.00  & 8.67  \\
      BIA (ViT-B/16) & 10.70 & 9.90  & 12.86 & 12.47 & 8.97  & 8.10  & 7.50  & 5.03  \\
      MI (ViT-B/16) & 7.81  & 7.92  & 11.62 & 12.96 & 8.26  & 7.51  & 6.46  & 6.96  \\
      PNA (ViT-B/16) & 7.13  & 8.58  & 10.79 & 14.06 & 8.03  & 7.98  & 6.11  & 7.71  \\
      TGR (ViT-B/16) & 12.73 & 11.55 & 16.18 & 18.34 & 12.16 & 11.50 & 8.88  & 9.32  \\
      ATT (ViT-B/16) & 12.22 & 12.05 & 17.70 & 19.19 & 12.04 & 11.27 & 8.65  & 10.49 \\
      \midrule
      \rowcolor{gray!15}
      \dsva{} (DINO) & \rb{20.88} & 19.47 & \rb{23.93} & \rb{26.28} & 21.49 & \rb{15.96} & \rb{12.80} & 11.67 \\
      \rowcolor{gray!15}
      \dsva{} (MAE) & 15.11 & 14.69 & 14.52 & 18.46 & 15.94 & 11.50 & 10.04 & 10.39 \\
      \rowcolor{gray!15}
      \dsva{} (Joint) & 19.19 & \rb{19.64} & 21.44 & 24.45 & \rb{22.31} & 14.79 & 12.11 & \rb{11.99} \\
      \bottomrule
  \end{tabular}}
  \caption{\textbf{Additional transferability comparisons against models with
    defenses.} We include additional comparisons in defense evasion against
    various robust ConvNets, ViTs, and hybrid models equipped with
  state-of-the-art adversarial defenses.}
  \label{tab:additional-transferability-defense-models}
\end{table*}

\Cref{tab:transfer-cross-domain-coarse-grained} showcases our findings on
coarse-grained classification domain transferability. With the target models in
CIFAR-10 and CIFAR-100 being VGG-like architectures, the BIA attack using a
VGG-19 surrogate model unsurprisingly yields superior results. Among the
\dsva{} variants, \dsva{} (Joint) with DINO and MAE at stride $s=8$ excels,
closely matching the baseline performance in these domains. In contrast, for the
SVHN and STL-10 domains, \dsva{} variants outperform the baseline, with
\dsva{} (DINO) surpassing \dsva{} (Joint) in SVHN due to DINO's sensitivity to
global shape and structure, which aligns with the focus of the SVHN domain on
\textit{house numbers} (digit classification). Interestingly, self-attention
exploitation in \dsva{} does not enhance performance in this coarse-grained
context.

Turning to fine-grained classification transferability in
\cref{tab:transfer-cross-domain-fine-grained}, \dsva{} (Joint) with active
self-attention exploitation leads in most scenarios, outperforming nearly all
baselines except when the target model is \textit{Res-50}. Notably, \dsva{}
(DINO) outperforms the otherwise dominant \dsva{} (Joint) variant in a specific
case: attacking the \textit{Stanford Cars} domain's \textit{SE-Res-101} model.

Aggregating the results, we conclude that \dsva{} (Joint) variant remains the
most robust attack overall for even most challenging cross-domain transfer
scenarios, with the self-attention exploitation proving beneficial in most
cases.

\section{Additional Comparisons of Transferability to Defense Models}
\label{sec:additional-transferability-defense-models}

In this section, we present additional comparisons on the transferability of
\dsva{} to robust ConvNets, ViTs, and hybrid models with state-of-the-art
defenses, which are lacking in prior work. We report the results in
\cref{tab:additional-transferability-defense-models}, where the citations
accompanying the model names refer to the respective state-of-the-art
adversarial defenses employed on the model itself. Note that we here use the
same experimental setups as in \cref{sec:experiments}, except for employing a
larger $\varepsilon=16$ constraint, otherwise the transferability across all
evaluated attacks would be too low to be comparable.

We observe that \dsva{} still consistently outperforms the baselines across all
models, averaging 17.04\% black-box transferability, even against the most
resilient defenses. \dsva{} (DINO) outperforms the joint variant in some cases,
indicating that the shape/structural features are more adversarially impactful
for robust models with smooth decision boundaries. These remarkable results once
again underscore the robustness and effectiveness of our \dsva{}.

\begin{figure*}[thbp]
  \centering
  \includegraphics[width=0.96\linewidth]{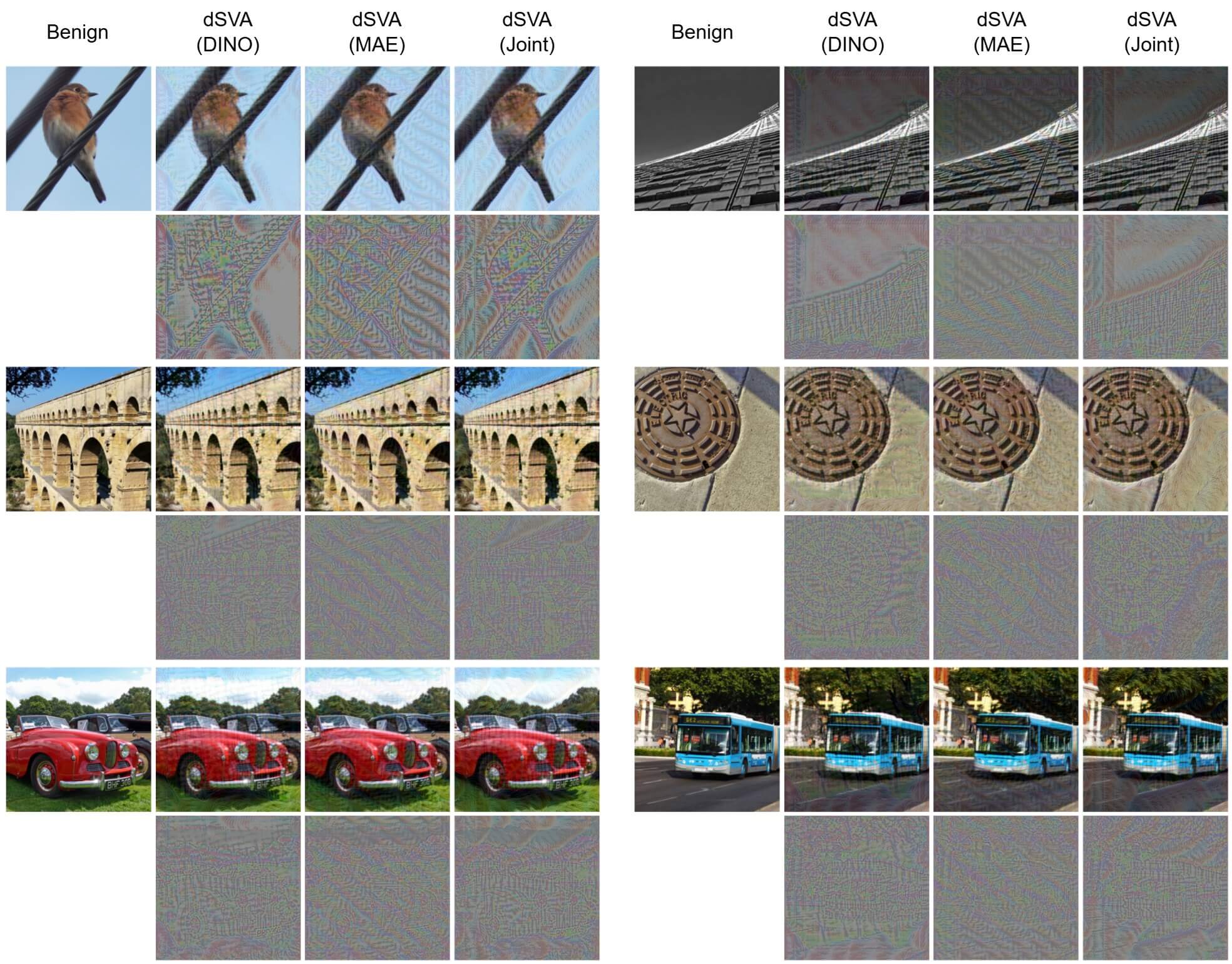}
  \caption{\textbf{Visualizations of adversarial examples.}
    We provide a few examples of side-by-side comparisons of the benign image,
    and adversarial examples generated by the 3 variants of \dsva{} (DINO, MAE,
  Joint). Perturbation is scaled and normalized for better visualization.}
  \label{fig:fig7}
\end{figure*}

\section{Visualization of Adversarial Examples}

In this section, we provide a few visual examples of the adversarial examples
and perturbations generated by \dsva{}. Figure \ref{fig:fig7} showcases several
instances of successful attacks by the 3 variants of \dsva{}: \dsva{} (DINO),
which emphasizes structural features; \dsva{} (MAE), which emphasizes textural
features; and \dsva{} (Joint), which successfully attends to both aspects, from
left to right respectively. These visualizations highlight the rich, impactful
perturbations crafted by our method, demonstrating its remarkable ability to
exploit model vulnerabilities effectively.

\section{Limitations and Future Work}

While \dsva{} demonstrates impressive black-box transferability by exploiting
self-supervised ViT features, we acknowledge certain limitations in our current
work and outline potential avenues for future work.

Although \dsva{} shows strong transferability in a digital settings, our current
work lacks full-scale physical world experiments. The potential of adopting
generative adversarial attacks for physical real-world scenarios is a complex,
challenging, yet valuable direction for future work.

Self-supervised methods with scaled training setups, such as DINOv2, may offer
potentially improved transferability for \dsva{}. Additionally, investigating
the use of ViTs with registers, and considering the use of multiple layers
during adversarial optimization, could further enhance the effectiveness and
robustness of \dsva{}. These approaches could lead to more effective adversarial
attacks and are crucial directions for future work.

We acknowledge the importance of ethical implications of our work, as with all
research in adversarial machine learning. Future research will continue to
explore the broader societal impacts of adversarial attacks and contribute to
the development of more robust and secure AI systems.